\def\paperID{}
\def\confName{CVPR}
\def\confYear{2026}

\def\paperTitle{Pose-Free Omnidirectional Gaussian Splatting for 360-Degree Videos with Consistent Depth Priors}

\def\authorBlock{
    Chuanqing Zhuang\textsuperscript{1} \qquad
    Xin Lu\textsuperscript{1} \qquad
    Zehui Deng\textsuperscript{1} \qquad
    Zhengda Lu\textsuperscript{1} \qquad  \\
    Yiqun Wang\textsuperscript{2} \qquad
    Junqi Diao\textsuperscript{3} \qquad
    Jun Xiao\textsuperscript{1$\dagger$} \\
    \textsuperscript{1}School of Artificial Intelligence,\quad University of Chinese Academy of Sciences \\
    \textsuperscript{2}Chongqing University \qquad \textsuperscript{3}Air Force Engineering University \qquad \textsuperscript{$\dagger$}\textit{Corresponding Author}\\
    {\tt\small \{zhuangchuanqing, \{luxin22, dengzehui23\}@mails, luzhengda, xiaojun\}@ucas.ac.cn} \\
    {\tt\small yiqun.wang@cqu.edu.cn, diaojunqi19@mails.ucas.edu.cn}
}

\newif\ifreview 
\newif\ifarxiv 
\newif\ifcamera \newcommand{\cameraready}{\cameratrue}
\newif\ifrebuttal 

\cameraready

\pdfoutput=1
\documentclass[10pt,twocolumn,letterpaper]{article}
\ifreview \usepackage[review]{cvpr} \fi
\ifarxiv \usepackage[pagenumbers]{cvpr} \fi
\ifrebuttal \usepackage[rebuttal]{cvpr} \fi
\ifcamera \usepackage{cvpr} \fi

\usepackage{graphicx}	
\usepackage{amsmath}	
\usepackage{amssymb}	
\usepackage{booktabs}
\usepackage{times}
\usepackage{microtype}
\usepackage{epsfig}
\usepackage{caption}
\usepackage{float}
\usepackage{placeins}
\usepackage{color, colortbl}
\usepackage{stfloats}
\usepackage{enumitem}
\usepackage{tabularx}
\usepackage{xstring}
\usepackage{multirow}
\usepackage{xspace}
\usepackage{url}
\usepackage{subcaption}
\usepackage{xcolor}
\usepackage[hang,flushmargin]{footmisc}
\usepackage{threeparttable}
\usepackage[T1]{fontenc}
\usepackage{newtxtext}
\usepackage{newtxmath}

\ifcamera \usepackage[accsupp]{axessibility} \fi

% ...

\ifarxiv  \fi

\newcommand{\R}[1]{{%
    \textbf{%
        \ifstrequal{#1}{1}{\textcolor{red}{R#1}}{%
        \ifstrequal{#1}{2}{\textcolor{blue}{R#1}}{%
        \ifstrequal{#1}{3}{\textcolor{magenta}{R#1}}{%
        \ifstrequal{#1}{4}{\textcolor{teal}{R#1}}{%
                           \textcolor{cyan}{R#1}%
        }}}}%
    }%
}}

\usepackage{xr-hyper}

\makeatletter
\def\cvprsubsubsection{\@startsection{subsubsection}{3}{0pt}{0.5em}{-0.5em}{\normalfont\normalsize\bfseries}}
\def\cvprssubsubsect#1{\cvprsubsubsection*{#1}}
\def\cvprsubsubsect#1{\cvprsubsubsection{#1}}
\def\subsubsection{\@ifstar\cvprssubsubsect\cvprsubsubsect}
\makeatother

\makeatletter
\newcommand*{\addFileDependency}[1]{
  \typeout{(#1)}
  \@addtofilelist{#1}
  \IfFileExists{#1}{}{\typeout{No file #1.}}
}

\makeatother
\newcommand*{\myexternaldocument}[1]{
    \externaldocument{#1}
    \addFileDependency{#1.tex}
    \addFileDependency{#1.aux}
}

\definecolor{cvprblue}{rgb}{0.21,0.49,0.74}
\usepackage[pagebackref,breaklinks,colorlinks,allcolors=cvprblue]{hyperref}
\usepackage[capitalize]{cleveref}
\crefname{section}{Sec.}{Secs.}
\crefname{table}{Table}{Tables}
\crefname{figure}{Fig.}{Figs.}

\ifarxiv \crefname{appendix}{App.}{Apps.}
\else \crefname{appendix}{Suppl.}{Suppls.} \fi

\unless\ifarxiv \myexternaldocument{_supplementary} \fi

\begin{document}
\title{\paperTitle}
\author{\authorBlock}

\twocolumn[{%
\renewcommand\twocolumn[1][]{#1}%
\maketitle
\includegraphics[width=\linewidth]{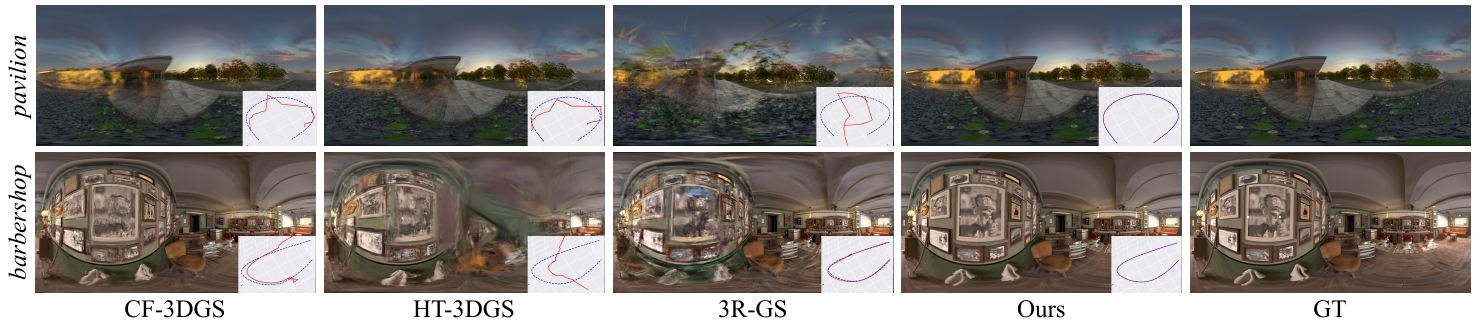}
\captionof{figure}{
\textbf{Example results of novel view synthesis and camera pose estimation compared with previous pose-free methods.}
In both the \textit{pavilion} and \textit{barbershop} scenes, our method accurately reconstructs camera poses and synthesizes realistic novel view panoramic images.
}
\label{fig:0-teaser}
\vspace{1em}
}]


\begin{abstract}
Omnidirectional 3D Gaussian Splatting with panoramas is a key technique for 3D scene representation, and existing methods typically rely on slow SfM to provide camera poses and sparse points priors.
In this work, we propose a pose-free omnidirectional 3DGS method, named PFGS360, that reconstructs 3D Gaussians from unposed omnidirectional videos.
To achieve accurate camera pose estimation, we first construct a spherical consistency-aware pose estimation module, which recovers poses by establishing consistent 2D-3D correspondences between the reconstructed Gaussians and the unposed images using Gaussians' internal depth priors.
Besides, to enhance the fidelity of novel view synthesis, we introduce a depth-inlier-aware densification module to extract depth inliers and Gaussian outliers with consistent monocular depth priors, enabling efficient Gaussian densification and achieving photorealistic novel view synthesis.
The experiments show significant outperformance over existing pose-free and pose-aware 3DGS methods on both real-world and synthetic 360-degree videos.
Code is available at \href{https://github.com/zcq15/PFGS360}{https://github.com/zcq15/PFGS360}.
\end{abstract}
\section{Introduction}
\label{sec:intro}

Omnidirectional 3D Gaussian Splatting (3DGS) for 360-degree videos has emerged as a pivotal technique for VR applications, including virtual tourism and interior visualization.
Although pose-free approaches without SfM are widely explored in perspective 3DGS, existing omnidirectional 3DGS methods \cite{lee2024odgs,huang2024error,li2025omnigs} still rely on time-consuming SfM priors \cite{adorjan2016opensfm,moulon2016openmvg,jiang20243d}.
This stems from instability in pose estimation during rasterization introduced by the spherical projection of panoramas, degrading pose estimation and rendering quality when perspective pose-free methods are directly applied to panoramas.
These challenges motivate the development of a pose-free 3DGS for 360-degree videos.

To recover camera poses without SfM, previous methods  \cite{fu2024colmap, ji2025sfm, yao2025smallgs} rasterize images from reconstructed Gaussians and back-propagate gradients through the rasterizer to optimize camera poses.
However, the Jacobian of the omnidirectional 3DGS \cite{lee2024odgs,li2025omnigs} significantly amplifies Gaussian errors near the poles, undermining the stability of pose gradients.
In addition, the spherical affine approximation of omnidirectional 3DGS introduces different projection errors across regions, further disturbing the pose gradient during rasterization, particularly among frames with large viewpoint variations.
Considering the instability of camera pose gradients during rasterization, we build upon classical PnP(Perspective-n-Point)-based pose solvers to achieve stable panoramic pose estimation.
Moreover, some recent perspective methods \cite{chen2024zerogs,cai2024dust,huang20253r} introduce correspondence priors based on 3D visual foundation models (VFMs) to enable pose-free 3DGS for perspective images, but VFMs trained for perspective images cannot directly process panoramas.

Furthermore, we observe that dense geometric priors using monocular depth estimation (MDE) models or 3D VFMs are critical for pose-free perspective 3DGS methods \cite{ji2025sfm, lin2025longsplat, cong2025videolifter} to outperform SfM-aware approaches.
By merging multi-frame point clouds into the Gaussians, they provide a rich set of seed points for Gaussian representations.
This strategy addresses the inherent limitation of sparse SfM initialization, which often fails to capture fine-grained scene details.
However, geometric priors remain inconsistent and inaccurate across different frames, which results in outliers within the Gaussian representation and hinders the construction of high-quality 3DGS.

To address them, we propose a pose-free omnidirectional 3DGS framework, PFGS360.
Specifically, we first introduce a spherical consistency–aware pose estimation module, which establishes 2D–3D correspondences between reconstructed Gaussians and unposed images, enabling accurate camera pose estimation via a spherical consistency–aware pose solver.
Next, we design a depth-inlier–aware densification module, which enhances Gaussian densification efficiency by aggregating geometrically consistent inliers from multi-frame depth priors and utilizing Gaussian outlier pruning to suppress the influence of depth noise.
Together, these components enable high-fidelity pose-free omnidirectional Gaussian splatting, supporting both photorealistic panoramic novel view synthesis and accurate camera pose estimation.
In summary, our main contributions are as follows:


\begin{itemize}
\item[{\scriptsize $\bullet$}] We propose a pose-free omnidirectional 3DGS framework that enables 3D Gaussian Splatting from unposed 360-degree videos, eliminating the dependence on SFM priors.
\item[{\scriptsize $\bullet$}] We introduce a spherical consistency–aware pose estimation module for accurate camera pose estimation and a depth-inlier–aware densification module for efficient Gaussian densification in 360-degree videos, improving pose estimation accuracy and rendering quality.
\item[{\scriptsize $\bullet$}] Experiments demonstrate that our method significantly outperforms existing approaches in both panoramic novel view synthesis and camera pose estimation for pose-free omnidirectional 3DGS.
\end{itemize}
\section{Related Work}
\label{sec:related}

\subsection{Omnidirectional 3DGS}

Building on the success of 3DGS \cite{kerbl20233d} for perspective images, subsequent works \cite{lee2024odgs,huang2024error,li2025omnigs} extend Gaussian rasterization to panoramic images using a spherical projection model.
This enables fully omnidirectional Gaussian rendering while avoiding the projection inconsistencies caused by perspective reprojection and stitching across different viewing directions \cite{radl2024stopthepop}, thus establishing the foundation for omnidirectional 3DGS applications.
Additionally, 360-GS \cite{bai2025360} removes the reliance on SfM priors in omnidirectional 3DGS through layout-based scene constraints. However, it is limited to indoor environments with specific spatial layouts.
SC-OmniGS \cite{huang2025sc} further incorporates calibration techniques to address camera distortion and SfM noise, enhancing omnidirectional 3DGS performance on real-world captured data.
Moreover, ray-based Gaussian rasterization methods \cite{blanc2025raygauss,moenne20243d,wu20253dgut,byrski2025raysplats} enable projection-model-independent Gaussian rendering, providing unified support across diverse camera types.
More recently, approaches such as \cite{chen2025splatter,zhang2025pansplat,lee2025omnisplat,ren2025panosplatt3r} integrate feed-forward 3DGS into omnidirectional images, achieving generalizable panoramic novel view synthesis.
Despite these advances, these methods incur high computational and memory costs and struggle to handle long-sequence panoramic videos efficiently.
In this work, we propose a pose-free omnidirectional 3DGS framework that eliminates the SfM priors by jointly integrating camera pose estimation and Gaussian Splatting, while aggregating multi-frame depth priors to further enhance rendering quality.

\subsection{Pose-Free 3DGS}

3DGS methods typically rely on SfM initialization to obtain sparse points and camera poses.
To eliminate this dependency, subsequent works \cite{fu2024colmap,ji2025sfm,liu2025adc,sun2024correspondence,dong2025towards} construct local monocular Gaussians using MDE models and optimize new frame camera poses through adjacent-frame rendering alignment.
However, errors in depth estimation accumulate in the relative camera poses, and Gaussian rasterization fails to provide correct optimization gradients when the camera undergoes large pose variations.
Moreover, the varying projection approximations across different regions in spherical rasterization substantially degrade the accuracy of relative pose estimation for omnidirectional monocular Gaussians.
Subsequent works \cite{chen2024zerogs,cai2024dust,gao2025easysplat,huang20253r,lin2025longsplat,fan2024instantsplat,paul2024gaussian} leverage 3D VFMs \cite{wang2024dust3r,leroy2024grounding,duisterhof2024mast3r,duisterhof2025mast3r,wang20253d} to achieve more robust pose estimation by exploiting 2D–3D correspondence priors.
The incorporation of dense point priors from VFMs notably enhances pose estimation accuracy and improves the fidelity of 3DGS representations.
Nevertheless, these foundation models are primarily designed for perspective images and fail to provide stable and accurate 2D–3D correspondences for panoramic inputs.
Additionally, the high resolution of 360-degree images introduces considerable computational and time overhead.
In this work, we reconstruct camera poses by directly establishing 2D–3D correspondences between the Gaussian model and the unposed images, without relying on any external VFM priors.
Furthermore, by extracting depth inliers and Gaussian outliers, we effectively filter unreliable geometric priors from the monocular depth estimation model, thereby achieving photorealistic, pose-free omnidirectional 3D Gaussian Splatting.
\section{Method}
\label{sec:method}

\begin{figure*}[t]
\centering
\includegraphics[width=0.85\linewidth]{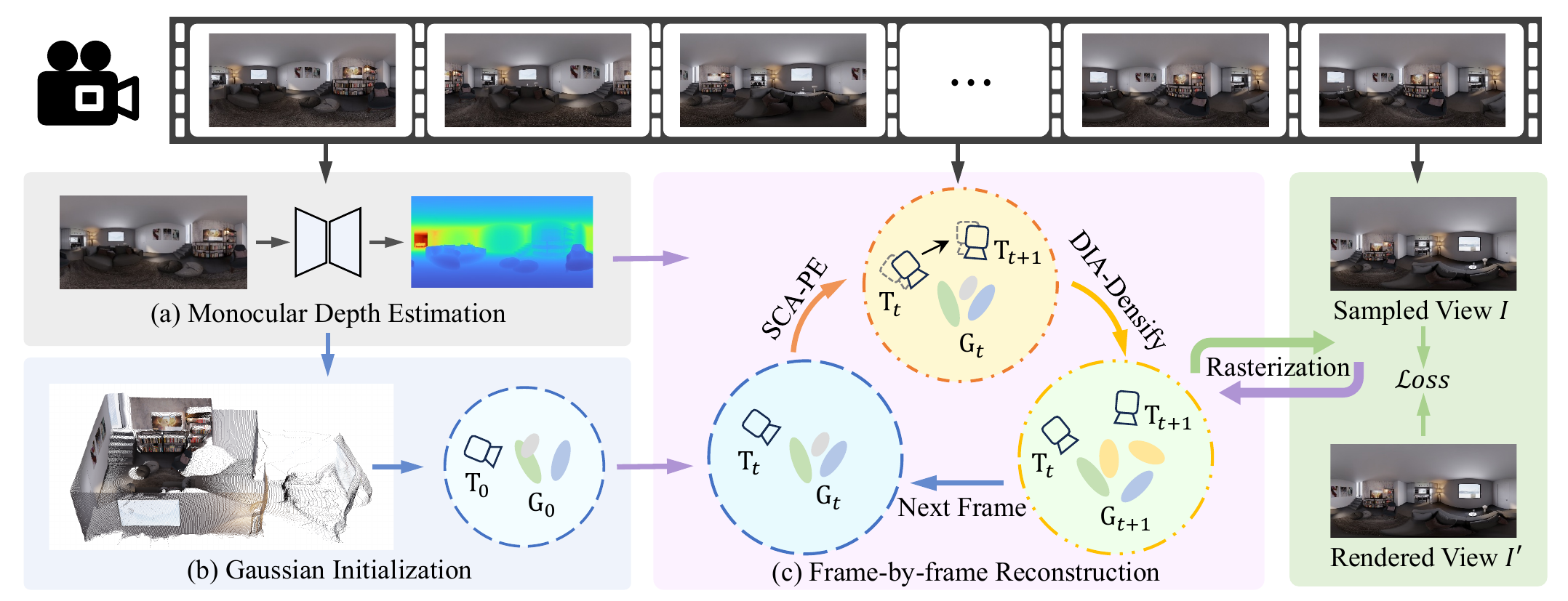}
\caption{
\textbf{An overview of our PFGS360 framework.}
(a) The PFGS360 model first predicts the monocular depth map of the initial panoramic frame.
(b) The omnidirectional 3DGS is then initialized using reprojected 3D points derived from the estimated depth.
(c) Subsequently, both the spherical consistency–aware pose estimation (SCA-PE, Section~\ref{sec:method-pose}) and depth-inlier–aware densification (DIA-Densify, Section~\ref{sec:method-gaussian}) are performed in a frame-by-frame manner to sequentially fuse panoramic frames and optimize the Gaussian \cite{kerbl20233d} representations.
}

\label{fig:3.1-overview}
\end{figure*}


We first provide an overview of the pose-free omnidirectional Gaussian Splatting (PFGS360) framework.
As illustrated in Figure~\ref{fig:3.1-overview}, given a sequence of unposed panoramic frames ${I_0, I_1, \cdots, I_{N}}$ from a 360-degree video, we initialize the Gaussians using the monocular depth map $D_0$ estimated from the first frame $I_0$.
For each subsequent frame $I_{t+1}$, we estimate its camera pose $T_{t+1}$ via the spherical consistency–aware pose estimation module (SCA-PE, Section~\ref{sec:method-pose}) and densify the current Gaussian model $G_t$ using the depth-inlier–aware densification module (DIA-Densify, Section~\ref{sec:method-gaussian}).
Finally, the Gaussians $G_t$ and all visited camera poses are jointly optimized and used to initialize $G_{t+1}$ for the next frame iteration.

\subsection{Spherical Consistency-aware Pose Estimation}\label{sec:method-pose}



To achieve accurate pose estimation, we employ a spherical consistency–aware (SCA) pose solver to obtain scale-consistent relative poses, followed by a post-optimization refinement to further improve camera pose accuracy.

\begin{figure}[t]
\center
  \includegraphics[width=\columnwidth]{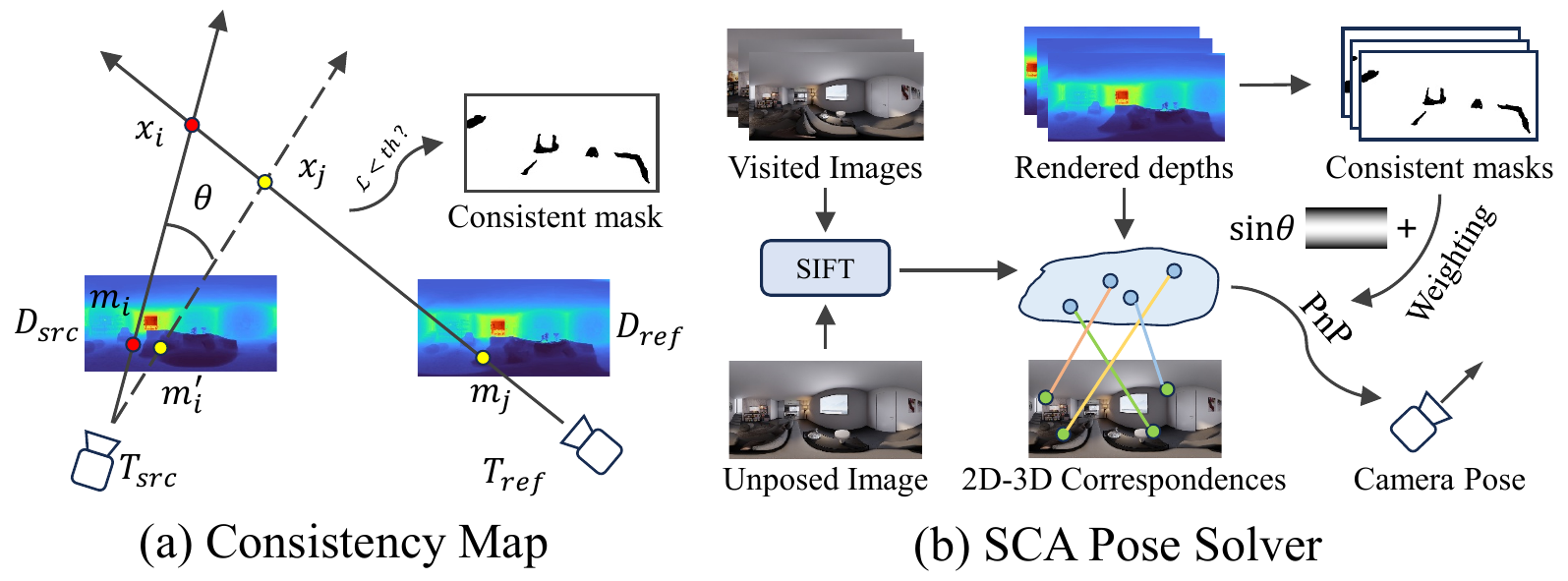}
  \caption{\textbf{Illustration of consistency map and SCA pose solver.}}
\label{fig:3.2-pose}
\end{figure}

\subsubsection*{Spherical Consistency-aware Pose Solver.}


To estimate the camera pose of a new frame $I_{t+1}$, we utilize the reconstructed Gaussians $G_t$ to establish 2D–3D correspondences between the previously visited and current unposed views, while introducing spherical consistency–awareness to mitigate the adverse effects of Gaussian reconstruction errors on pose estimation.
Specifically, we first render a depth map $D_r^k$ for each previously visited view $I_k$, providing scale-consistent geometric priors across frames.
However, since 3DGS relies on photometric optimization, the model often overfits to scene appearance, resulting in local geometric inaccuracies.
To reduce the impact of these depth errors on pose estimation, we compute spherical reprojection consistency between the rendered depth maps of different frames.

As illustrated in Figure~\ref{fig:3.2-pose}(a), for each pixel $\mathbf{m}_i$ in the source image $I_{src}$, we project it onto the reference image $I_{ref}$ using the corresponding camera poses and depth maps to obtain its projected pixel $\mathbf{m}_j$.
The pixel $\mathbf{m}_j$ is then reprojected back to the source image as $\mathbf{m}'_i$.
The spherical reprojection consistency between the two depth maps, $D_{src}$ and $D_{ref}$, is subsequently evaluated using the reprojected tangent angular error \cite{pagani2011structure} and the depth error:
\begin{equation}\label{eq:adj_mask}
\begin{aligned}
C_{src,ref}(\mathbf{x}) &= \mathcal{I}(\mathcal{L}_{tan}(\mathbf{m}_i,\mathbf{m}'_i) \le \epsilon_{tan}) \\
& \cap  \mathcal{I}( \text{abs}(\|\mathbf{x}_i\|- \|\mathbf{x}'_i\|)/\|\mathbf{x}_i\|\le \epsilon_{dep})
\end{aligned}
\end{equation}
where $\mathcal{I}(\cdot)$ is the indicator function, $\epsilon_{tan}=0.008$ and $\epsilon_{dep}=0.05$.

Next, as shown in Figure~\ref{fig:3.2-pose}(b), for each visited frame $I_k$, we compute the cross-frame consistency map of the rendered depth maps as
$M_{con}^k = C_{k,t} \times C_{k,k-1}$,
which jointly captures the local consistency between adjacent frames and the global consistency across multiple frames.
We then extract feature correspondences between the visited frames ${I_0, I_1, \cdots, I_t}$ and the unposed new frame $I_{t+1}$ using SIFT descriptors, as SIFT has demonstrated strong robustness in panoramic image matching \cite{jiang20243d, ren2025panosplatt3r}.
With the rendered depth maps ${D_r^0, D_r^1, \cdots, D_r^t}$ and their corresponding consistency masks ${M_{con}^0, M_{con}^1, \cdots, M_{con}^t}$, we obtain a set of 2D–3D correspondences $(\mathbf{x}_k, \mathbf{m}_{t+1})$.
Finally, the camera pose $T_{t+1}$ for the new frame $I_{t+1}$ is estimated by minimizing the spherical consistency–aware reprojection error, defined as:
\begin{gather}
T = \arg \min \sum \lambda  \mathcal{L}_{tan}(\mathbf{x}_k,\mathbf{m}_{t+1}) \\
\lambda  = \sin(\mathbf{m}'_{k}) \sin(\mathbf{m}_{t+1}) M_{con}^k(\mathbf{x}_k) \\
\mathcal{L}_{tan}(\mathbf{x}_k,\mathbf{m}_{t+1}) = 2\sqrt{ \frac {1-\mathbf{m}'_{k}\mathbf{m}_{t+1}}  {1+\mathbf{m}'_{k}\mathbf{m}_{t+1}}}
\label{eq:tan-err}
\end{gather}
Here, $\mathbf{m}'_{k}$ denotes the reprojected pixel of $\mathbf{x}_k$ in the camera coordinates of $I_{t+1}$, and the coefficient $\lambda$ introduces spherical balancing \cite{zioulis2019spherical} while enforcing cross-frame consistency.

\subsubsection*{Post Pose Refinement.}

After obtaining an initial coarse camera pose using the pose solver, we refine the poses of the visited cameras.
We find that directly optimizing camera poses with a photometric loss while keeping Gaussian parameters fixed is suboptimal.
Geometric inaccuracies in the reconstructed Gaussians and missing regions in newly observed frames introduce noise into the photometric loss, degrading pose optimization accuracy.
To mitigate this issue, we apply an adjacent-frame consistency mask
$M_{adj}^k=C_{k,k-1}\times C_{k,k+1}$
as a weighting term in the loss function, enhancing robustness during pose refinement, i.e.,
\begin{equation}
\begin{aligned}
\mathcal{L}_{photo} = &(1-\lambda_{dssim}) M_{adj} \mathcal{L}_{l1}(I',I) \\ &+\lambda_{dssim} M_{adj} \mathcal{L}_{dssim}(I',I)
\end{aligned}
\end{equation}
where $\lambda_{dssim}=0.2$.


\begin{figure}[t]
\center
  \includegraphics[width=\columnwidth]{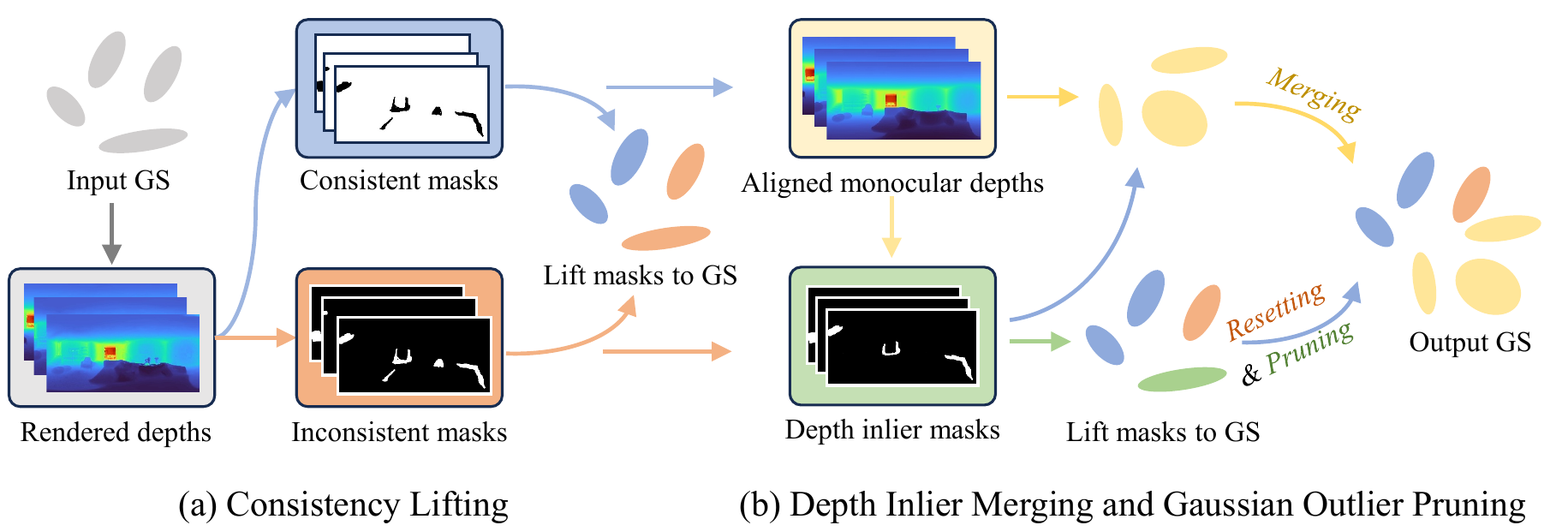}
  \caption{\textbf{Illustration of DIA-Densify module.}}
\label{fig:3.3-gaussian}
\end{figure}

\subsection{Depth-inlier-aware Densification}\label{sec:method-gaussian}




To reconstruct higher-fidelity Gaussians, we propose a depth-inlier–aware densification (DIA-Densify) module, which leverages geometrically consistent depth priors to refine the Gaussians by aggregating depth inliers and pruning Gaussian outliers.
Note, we extract inliers and outliers from the first $t$ frames, since the coarse pose $T_{t+1}$ is insufficiently accurate.

\subsubsection*{Depth Inlier Merging.}

We first introduce a depth-inlier merging module, which selects high-confidence depth points based on geometric consistency and patch-level similarity, and integrates them into the Gaussian model as inliers.
As shown in Figure~\ref{fig:3.3-gaussian}(a), for each rendered depth map $D_r^k$, we first compute the consistent and inconsistent regions, $M_{con}^k$ and $M_{inc}^k$, respectively.
These regions are lifted to the Gaussians $G_t$ to identify inliers and outliers in $G_t$ (see \textbf{Sec. Gaussian Outlier Pruning}).
Next, as shown in Figure~\ref{fig:3.3-gaussian}(b), we align the monocular depth $D_m^k$ to the current 3DGS scale using the consistency mask $M_{con}^k$, following \cite{ranftl2020towards}:
\begin{gather}
D_{a}^k = \lambda_s D_{m}^k + \lambda_t \\
\lambda_s,\lambda_t = \arg\min \sum M_{con}^k\cdot (\lambda_s D_{m}^k + \lambda_t- D_{r}^k)
\end{gather}
and extract the region $M_{con,a}^k$ that satisfies adjacent-frame consistency.
Despite this alignment, $M_{con,a}^k$ often contains unreliable depth estimates.
For instance, depth estimation frequently misses fine geometric details and produces over-smoothed surfaces across multiple frames, which cannot be fully corrected by multi-view geometric consistency alone.
Motivated by patch-based multi-view stereo methods \cite{furukawa2009accurate}, we employ a normalized cross-correlation (NCC) similarity metric to identify high-confidence depth inliers:
\begin{equation}
\begin{aligned}
M_{ncc}^k = & \mathcal{I}(\mathcal{F}(I_k,\mathcal{W}(I_{k-1},D_{a}^k))>\mathcal{F}(I_k,\mathcal{W}(I_{k-1},D_{r}^k)))
\end{aligned}
\end{equation}
where $\mathcal{W}(I_{ref},D_{src})$ denotes warping the reference image $I_{ref}$ into the viewpoint of $I_{src}$ using depth $D_{src}$, and $\mathcal{F}(\cdot, \cdot)$ represents the NCC function.
Finally, the reprojected 3D points from the candidate regions
$M_{inlier}^k =  M_{inc}^k \cap  M_{con,a}^k \cap M_{ncc}^k$
across all visited frames are merged into the Gaussian model $G_t$, enriching the reconstructed geometry with reliable depth inliers, as shown in Figure~\ref{fig:3.3-gaussian}(b).

\subsubsection*{Gaussian Outlier Pruning.}\label{sec:3.x.x-outlier}

To further reduce reconstruction errors, we introduce a Gaussian outlier pruning module, which removes Gaussians that have been replaced by depth inliers and resets the opacity of the remaining Gaussian outliers, as shown in Figure~\ref{fig:3.3-gaussian}(b).
Specifically, using the rendered depth consistency masks $M_{con}^k$, inconsistency masks $M_{inc}^k$, and the aligned depth inlier masks $M_{inlier}^k$, we accumulate these masks onto the Gaussian $G_t$ as follows:
\begin{align}
A_i = \frac{\sum_{j\in \mathcal{I}_{i}^{max}} \omega_{i,j} M_j}{\sum_{j\in \mathcal{I}_{i}^{max}} \omega_{i,j}}
\end{align}
Here, $M_j$ denotes the value of $M_{con}^k$ or $M_{inc}^k$ at pixel $j$, and $A_i$ represents the corresponding accumulated result assigned to $i$-th Gaussian $G_{t}^i$.
$\mathcal{I}_{i}^{max}$ is the set of pixel indices across all visited images $\{I_0,I_1,\cdots,I_t\}$ where the Gaussian $G_{t}^i$ exhibits the maximum rendering weight $\omega$, following \cite{fang2024mini}.
Accordingly, all Gaussians with $A_{inlier} > 0.8$ are replaced by depth inliers and pruned from the model, whereas those satisfying $A_{inlier} \le 0.8  \ \cap A_{inc} > 0.8$ are reset to their initial opacities.
Finally, we perform joint optimization of Gaussians and camera poses, and the optimized Gaussians serve as $G_{t+1}$ for integrating the subsequent frame $I_{t+1}$ in the next iteration cycle.


\section{Experiments}
\label{sec:exp}


In this section, we present experiments to validate the effectiveness of the proposed method.
We first describe the experimental setup and the datasets used.
Next, we evaluate the method on NVS and camera pose estimation tasks using these datasets.
Finally, we conduct ablation studies to analyze the contribution of each component in our framework.

\begin{table*}[ht]
\caption{\textbf{Quantitative comparison results for novel view synthesis on OB3D.}}
\label{tab:4.1-eval-nvs-ob3d}
\centering
\scriptsize
\setlength{\tabcolsep}{2pt}
\begin{threeparttable}
\resizebox{0.9\textwidth}{!}{%
\begin{tabular}{ccccccccccccccccccc}
\toprule
\multirow{2}{*}{Method} 
  & \multicolumn{9}{c}{OB3D-Egocentric} 
  & \multicolumn{9}{c}{OB3D-NonEgocentric} \\
\cmidrule(lr){2-10}\cmidrule(lr){11-19}
  & \multicolumn{3}{c}{indoor} 
  & \multicolumn{3}{c}{outdoor} 
  & \multicolumn{3}{c}{mean}
  & \multicolumn{3}{c}{indoor} 
  & \multicolumn{3}{c}{outdoor} 
  & \multicolumn{3}{c}{mean} \\
\cmidrule(lr){2-4}\cmidrule(lr){5-7}\cmidrule(lr){8-10}
\cmidrule(lr){11-13}\cmidrule(lr){14-16}\cmidrule(lr){17-19}
  & PSNR & SSIM & LPIPS
  & PSNR & SSIM & LPIPS
  & PSNR & SSIM & LPIPS
  & PSNR & SSIM & LPIPS
  & PSNR & SSIM & LPIPS
  & PSNR & SSIM & LPIPS \\
\midrule
ODGS    & 28.14 & 0.840 & 0.241 & 27.47 & 0.849 & 0.176 & 27.76 & 0.845 & 0.204 & 26.80 & 0.819 & 0.256 & 25.04 & 0.771 & 0.243 & 25.79 & 0.792 & 0.249 \\
OmniGS  & \underline{\textit{33.25}} & \underline{\textit{0.897}} & 0.169 & 29.89 & \underline{\textit{0.907}} & 0.112 & \underline{\textit{31.35}} & \underline{\textit{0.903}} & 0.137 & 27.93 & 0.839 & 0.247 & \underline{\textit{25.86}} & \underline{\textit{0.785}} & \underline{\textit{0.258}} & \underline{\textit{26.75}} & \underline{\textit{0.808}} & 0.253 \\
\cmidrule(lr){1-19}
CF-3DGS & 25.14 & 0.737 & 0.314 & 25.14 & 0.745 & 0.208 & 25.14 & 0.742 & 0.253 & 22.13 & 0.684 & 0.385 & 22.44 & 0.670 & 0.291 & 22.31 & 0.676 & 0.330 \\
HT-3DGS & 26.24 & 0.753 & 0.276 & 24.78 & 0.731 & 0.218 & 25.39 & 0.740 & 0.242 & 22.92 & 0.694 & 0.351 & 22.30 & 0.669 & 0.288 & 22.56 & 0.680 & 0.314 \\


3R-GS & 31.87 & 0.892 & \underline{\textit{0.105}}  & \underline{\textit{30.63}} & 0.905 & \underline{\textit{0.073}} & 31.15 & 0.899 & \underline{\textit{0.086}} & \underline{\textit{28.06}} & \underline{\textit{0.841}} & \underline{\textit{0.137}} & 20.05 & 0.664 & 0.344 & 23.39 & 0.681 & \underline{\textit{0.238}} \\
Ours & \textbf{36.03} & \textbf{0.935} & \textbf{0.074} & \textbf{35.58} &\textbf{ 0.954} & \textbf{0.045} & \textbf{35.77} & \textbf{0.946 }&\textbf{ 0.057} & \textbf{32.40} & \textbf{0.908} & \textbf{0.113} &\textbf{ 29.68} & \textbf{0.872} &\textbf{ 0.114 }& \textbf{30.81} & \textbf{0.887} & \textbf{0.113} \\

\bottomrule
\end{tabular}%
}
\begin{tablenotes}
\footnotesize
\item \textit{Note.} Higher PSNR and SSIM indicate better reconstruction quality, while lower LPIPS indicates better perceptual similarity.
\end{tablenotes}
\end{threeparttable}
\end{table*}

\begin{table*}[ht]
\caption{\textbf{Quantitative comparison results for camera pose estimation on OB3D.}}
\label{tab:4.3-eval-pose-ob3d}
\centering
\scriptsize
\setlength{\tabcolsep}{2pt}
\begin{threeparttable}
\begin{tabular}{ccccccccccccccccccc}
\toprule
\multirow{2}{*}{Method} 
  & \multicolumn{9}{c}{OB3D-Egocentric} 
  & \multicolumn{9}{c}{OB3D-NonEgocentric} \\
\cmidrule(lr){2-10}\cmidrule(lr){11-19}
  & \multicolumn{3}{c}{indoor} 
  & \multicolumn{3}{c}{outdoor} 
  & \multicolumn{3}{c}{all}
  & \multicolumn{3}{c}{indoor} 
  & \multicolumn{3}{c}{outdoor} 
  & \multicolumn{3}{c}{all} \\
\cmidrule(lr){2-4}\cmidrule(lr){5-7}\cmidrule(lr){8-10}
\cmidrule(lr){11-13}\cmidrule(lr){14-16}\cmidrule(lr){17-19}
  & RPE\_t & RPE\_r & ATE
  & RPE\_t & RPE\_r & ATE
  & RPE\_t & RPE\_r & ATE
  & RPE\_t & RPE\_r & ATE
  & RPE\_t & RPE\_r & ATE
  & RPE\_t & RPE\_r & ATE \\
\midrule
CF-3DGS 
  & 3.904 & 14.351 & 0.0720
  & 2.665 & 0.971 & 0.0554
  & 3.181 & 6.546 & 0.0623
  & 1.752 & 12.540 & 0.0726
  & \underline{\textit{1.856}} & \underline{\textit{1.097}} & \underline{\textit{0.0443}}
  & 1.813 & 5.865 & 0.0561 \\
HT-3DGS 
  & 3.825 & 16.671 & 0.0696
  & 3.053 & 1.877 & 0.0639
  & 3.375 & 8.041 & 0.0663
  & 2.310 & 10.832 & 0.0828
  & 2.120 & 2.337 & 0.0580
  & 2.199 & 5.876 & 0.0683 \\
3R-GS 
  & \underline{\textit{0.363}} & \underline{\textit{0.132}} & \underline{\textit{0.0035}}
  & \underline{\textit{0.392}} & \underline{\textit{0.129}} & \underline{\textit{0.0038}}
  & \underline{\textit{0.380}} & \underline{\textit{0.130}} & \underline{\textit{0.0036}}
  & \underline{\textit{0.387}} & \underline{\textit{0.150}} & \underline{\textit{0.0088}}
  & 2.143 & 2.719 & 0.0753
  & \underline{\textit{1.303}} & \underline{\textit{1.522}} & \underline{\textit{0.0439}} \\
Ours 
  & \textbf{0.018} & \textbf{0.015} & \textbf{0.0001}
  & \textbf{0.019} & \textbf{0.013} & \textbf{0.0003}
  & \textbf{0.018} & \textbf{0.014} & \textbf{0.0002}
  &\textbf{ 0.010 }&\textbf{ 0.016} & \textbf{0.0001}
  & \textbf{0.061} & \textbf{0.016} & \textbf{0.0011}
  & \textbf{0.040} &\textbf{ 0.016} & \textbf{0.0007} \\
\bottomrule
\end{tabular}
\begin{tablenotes}
\footnotesize
\item \textit{Note.} Lower RPE\_t, RPE\_r, and ATE values indicate higher pose estimation accuracy.
\end{tablenotes}
\end{threeparttable}
\end{table*}

\subsection{Experimental Implementations and Datasets}

The proposed PFGS360 method is implemented using PyTorch 
and Nerfstudio \cite{tancik2023nerfstudio} frameworks, with panoramic 3DGS rasterization developed via gsplat \cite{ye2025gsplat}.
For a fair comparison with pose-free 3DGS methods originally designed for perspective images, we adapt their pipelines by replacing the perspective MDE model with a panoramic MDE model \cite{piccinelli2025unik3d} and substituting the original perspective rasterization with our panoramic rasterization.
All experiments are conducted on a single NVIDIA RTX 4090 GPU.

We evaluate our method on the OB3D \cite{ito2025ob3d} and Ricoh360 \cite{choi2023balanced} datasets.
OB3D is a synthetic dataset comprising both indoor and outdoor scenes, with ground-truth camera poses exported from the rendering engine, enabling quantitative evaluation of novel view synthesis and pose estimation accuracy.
Ricoh360 contains panoramic videos captured in challenging real-world environments, which we use to assess the effectiveness of our method in practical scenarios.


\begin{figure*}[t]
\centering
\includegraphics[width=0.875\linewidth]{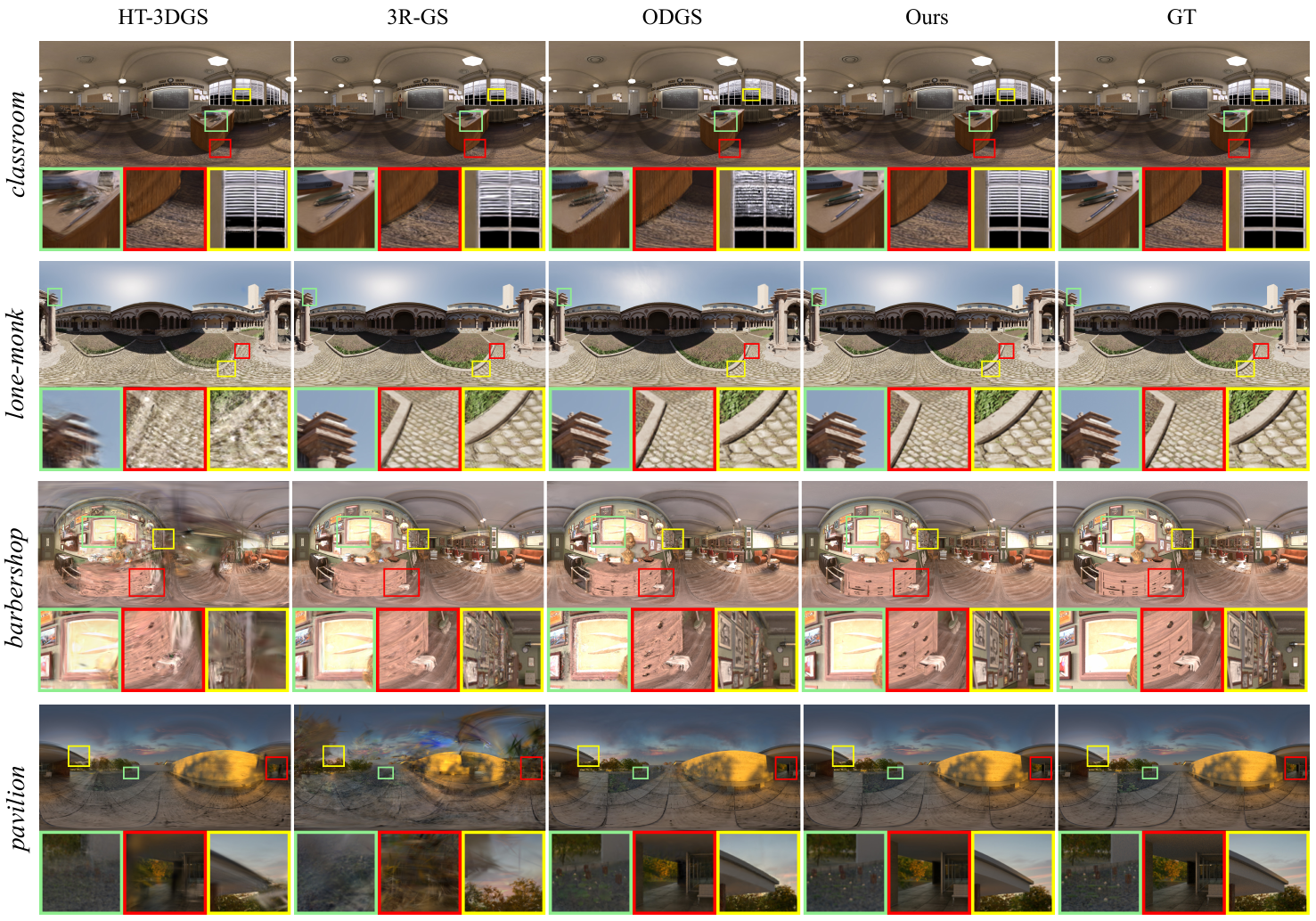}
\caption{\textbf{Visualization results for novel view synthesis on OB3D.}}

\label{fig:4.1-eval-ob3d-nvs}
\end{figure*}

\begin{figure}[t]
\centering
\includegraphics[width=\columnwidth]{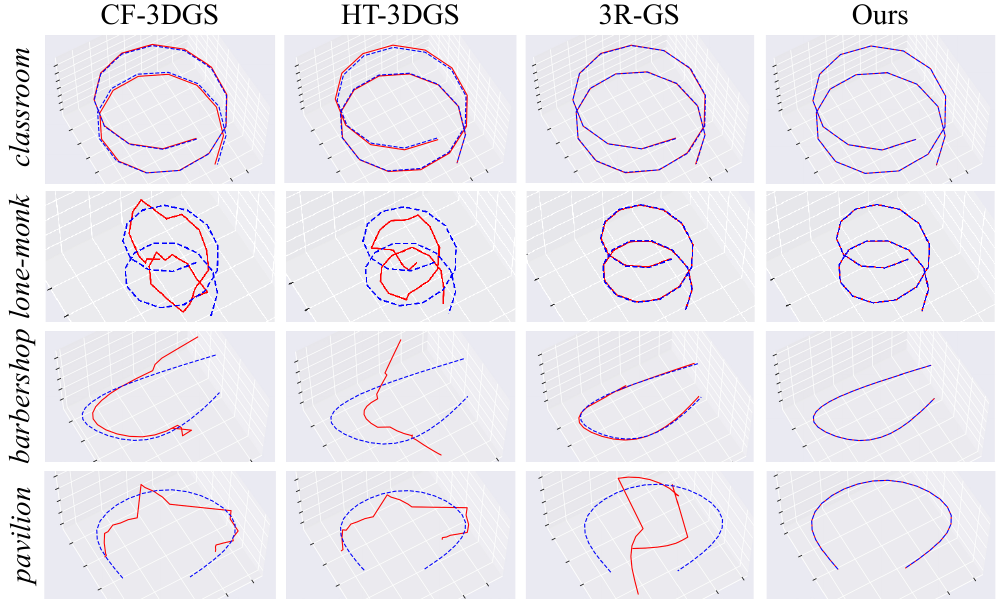}
\caption{\textbf{Visualization results for pose estimation on OB3D.}}

\label{fig:4.2-eval-ob3d-pose}
\end{figure}

\subsection{Evaluation on OB3D Dataset}
We begin by comparing our approach with existing novel view synthesis methods on the OB3D dataset, including both pose-free and pose-aware approaches.

\subsubsection*{Quantitative Comparison for Novel View Synthesis.}

We first perform a comparative evaluation on the OB3D dataset for novel view synthesis, considering both pose-free and pose-aware methods, as summarized in Table~\ref{tab:4.1-eval-nvs-ob3d}.
The pose-free methods CF-3DGS\cite{fu2024colmap} and HT-3DGS\cite{ji2025sfm} perform significantly worse than pose-aware approaches ODGS and OmniGS, due to unreliable camera pose estimation when adapted to panoramic images.
3R-GS\cite{huang20253r} achieves competitive results compared to pose-aware methods in the \textit{Egocentric} subset and indoor scenes of the \textit{NonEgocentric} subset, but its performance degrades notably in the more challenging outdoor scenes of the \textit{NonEgocentric} subset.
In contrast, our method leverages a consistency-aware pose estimation module to obtain accurate camera poses, achieving the best performance across all scenes and camera trajectories, and significantly outperforming pose-free methods.
Moreover, it surpasses pose-aware methods ODGS\cite{lee2024odgs} and OmniGS\cite{li2025omnigs}, which rely on ground-truth camera pose priors.
Specifically, our method improves the second-best results by 4.42 dB PSNR for \textit{Egocentric} trajectories and 4.06 dB PSNR for \textit{NonEgocentric} trajectories.
This performance gain is attributed to the aggregation of depth priors, which enables efficient Gaussian densification, particularly on the \textit{OB3D-NonEgocentric} dataset, where camera motion involves large viewpoint changes and sparse observations.


\subsubsection*{Quantitative Comparison for Pose Estimation.}\label{sec:exp-eval-pose}

Next, we evaluate the camera pose estimation accuracy of different methods on the OB3D dataset, as reported in Table~\ref{tab:4.3-eval-pose-ob3d}.
Both CF-3DGS and HT-3DGS reconstruct camera poses using similar adjacent-frame monocular Gaussian rendering strategies, which suffer from error accumulation over video sequences.
Furthermore, the spherical projection of panoramic images introduces non-affine distortions, so even small errors in monocular depth estimation can cause substantial distortions in the rendered Gaussians across adjacent frames.
This significantly increases the difficulty of optimizing camera poses via photometric loss, further degrading pose estimation accuracy.
3R-GS mitigates some of these issues by reprojecting panoramic images into overlapping perspective tiles and using MASt3R-SfM \cite{duisterhof2025mast3r} to reconstruct camera poses and sparse points with correspondences.
It achieves notable improvements over CF-3DGS and HT-3DGS in the \textit{OB3D Egocentric} subset with small camera motions.
However, it fails to recover accurate poses in the \textit{OB3D NonEgocentric} subset, where camera movements are larger.
In contrast, our method consistently reconstructs accurate camera poses across all trajectories, achieving an RPE\_t of 0.018 for \textit{Egocentric} and 0.040 for \textit{NonEgocentric} sequences.

\subsubsection*{Visual Comparison Results.}

Finally, we present a visual comparison of novel view synthesis and camera pose estimation results on the OB3D dataset.
As shown in the first two rows of Figures~\ref{fig:4.1-eval-ob3d-nvs} and \ref{fig:4.2-eval-ob3d-pose}, in the \textit{OB3D Egocentric} subset, HT-3DGS exhibits visually noticeable pose reconstruction errors, leading to prominent artifacts such as the pencil in \textit{classroom} and the pillar, floor tiles, and curbstones in \textit{lone-monk}.
3R-GS substantially improves pose accuracy but does not fully eliminate artifacts, which remain visible along object boundaries, including the pencil and blinds in \textit{classroom}, and the pillar and curbstones in \textit{lone-monk}.
In contrast, our method reconstructs camera poses that are nearly perfectly aligned with the ground truth, significantly reducing rendering artifacts.
It even surpasses ODGS in challenging regions, such as the blinds and curbstones, despite ODGS leveraging ground-truth poses. (For more detailed visualizations of the error maps, please refer to the \textbf{Supplementary Materials}.)

Moreover, as shown in the last two rows of Figures~\ref{fig:4.1-eval-ob3d-nvs} and \ref{fig:4.2-eval-ob3d-pose}, the larger camera displacements in the \textit{OB3D NonEgocentric} subset substantially increase the difficulty of camera pose estimation. 
In the \textit{barbershop} scene, HT-3DGS fails to reconstruct valid camera poses, while 3R-GS produces noticeable pose errors, resulting in artifacts and blurring around the cabinet and wall paintings.
In the more open \textit{pavilion} outdoor scene, 3R-GS fails entirely in pose estimation due to the scarcity of reliable correspondences.
In contrast, our method accurately reconstructs camera poses and renders high-quality novel panoramic views for both scenes.
It also outperforms the pose-aware ODGS in reconstruction fidelity, producing sharper textures for the cabinet and wall paintings in \textit{barbershop}, as well as finer cloud patterns and floor gap details in \textit{pavilion}.

\begin{figure*}[t]
\centering
\includegraphics[width=0.85\linewidth]{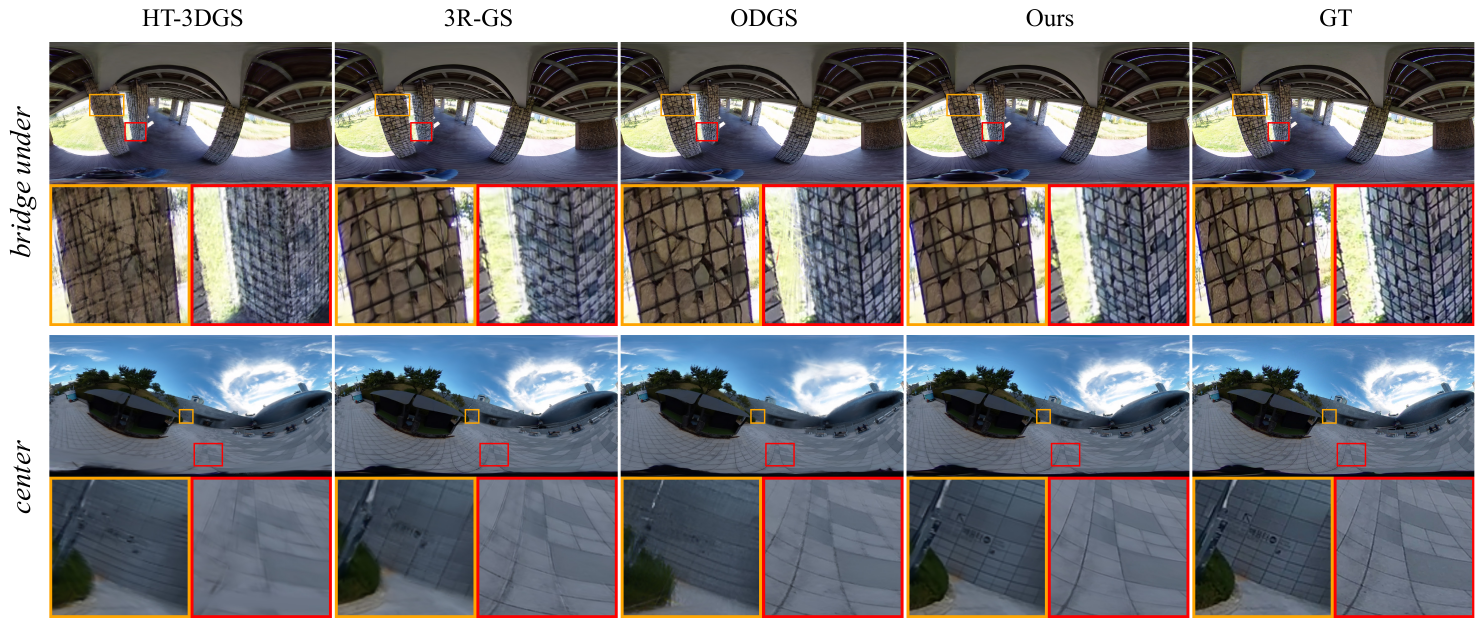}
\caption{\textbf{Visualization results for novel view synthesis on Ricoh360.}}

\label{fig:4.3-eval-ricoh-nvs}
\end{figure*}

\subsection{Evaluation on Ricoh360 Dataset}


\begin{table}[t]
\caption{\textbf{Quantitative comparison results for NVS on Ricoh360.}}
\label{tab:4.2-eval-nvs-ricoh360}
\centering
\scriptsize
\setlength{\tabcolsep}{3pt} 
\begin{threeparttable}
\begin{tabular}{ccccccc}
\toprule
Metric 
& {\rotatebox{45}{\scriptsize ODGS} }
& {\rotatebox{45}{\scriptsize OmniGS}} 
& {\rotatebox{45}{\scriptsize CF-3DGS}} 
& {\rotatebox{45}{\scriptsize HT-3DGS}} 
& {\rotatebox{45}{\scriptsize 3R-GS}} 
& {\rotatebox{45}{\scriptsize Ours}} \\
\midrule
PSNR & \underline{\textit{26.27}} & 26.03 & 22.77 & 23.00 & 24.18 & \textbf{28.05} \\
SSIM & \underline{\textit{0.846}} & 0.825 & 0.724 & 0.734 & 0.764 & \textbf{0.867} \\
LPIPS & \textbf{0.105} & \underline{\textit{0.128}} & 0.284 & 0.178 & 0.180 & 0.134 \\
\bottomrule
\end{tabular}
\end{threeparttable}
\end{table}

The Ricoh360 dataset presents several real-world challenges during data acquisition, including imaging distortions, motion blur, overexposure, and human occlusions.
Despite these adverse conditions, our method achieves the best overall performance, as reported in Table~\ref{tab:4.2-eval-nvs-ricoh360}, surpassing the second-best approach by 1.78 dB PSNR.
This improvement is attributed to its ability to leverage geometrically consistent depth cues, which enhance pose estimation accuracy and stability while reconstructing more realistic Gaussians.
Furthermore, as shown in Figure~\ref{fig:4.3-eval-ricoh-nvs}, our rendered results exhibit more realistic textures compared to other methods, for example, the stone pillars in the \textit{bridge under} scene and the wall and floor tiles in the \textit{center} scene.
These results demonstrate the robustness and superior performance of our method in real-world scenarios.

\subsection{Experiments for Ablation Study}

We perform ablation studies on the \textit{OB3D NonEgocentric} subset to analyze the contributions of the main modules in our framework and evaluate performance under different monocular depth prediction models.

\subsubsection*{Ablation Study on Model Components.}

We first analyze the contribution of each module.
Specifically, we construct a baseline model based on CF-3DGS, where the local monocular Gaussians used for adjacent-frame pose estimation are replaced with the globally reconstructed Gaussian model at the current frame $t$.
During subsequent optimization, we perform joint optimization of camera poses and Gaussians.
Building upon this baseline, we progressively integrate the consistency-aware pose estimation module and the depth-inlier–guided densification module to form the complete PFGS360 framework.

\begin{table}[t]
\caption{\textbf{Ablation experiments on model components.}}
\label{tab:4.4-abs-modules}
\centering
\scriptsize
\setlength{\tabcolsep}{3pt} 
\begin{threeparttable}
\begin{tabular}{ccccccc}
\toprule
Method & PSNR & SSIM & LPIPS & RPE\_t & RPE\_r & ATE \\
\midrule
\textit{B}                                  & 20.17 & 0.559 & 0.413 & 3.684 & 5.601 & 0.1217 \\
\textit{B}+PnP\cite{lin2025longsplat}       & 27.50 & 0.815 & 0.164 & 0.229 & 0.156 & 0.0076 \\
\textit{B+SCA-PE}                              & 28.99 & 0.858 & 0.133 & 0.077 & 0.027 & 0.0021 \\
\textit{B+SCA-PE+DIM}                            & \underline{\textit{29.70}} & \underline{\textit{0.872}} &\underline{\textit{ 0.122}} & \underline{\textit{0.057}} & \underline{\textit{0.025}} & \underline{\textit{0.0013}} \\
\textit{B+SCA-PE+DIM+GOP}                          & \textbf{30.81} & \textbf{0.886} & \textbf{0.113} & \textbf{0.040} & \textbf{0.016} & \textbf{0.0007} \\
\bottomrule
\end{tabular}
\end{threeparttable}
\end{table}

\begin{figure}[t]
\centering
\includegraphics[width=0.9\linewidth]{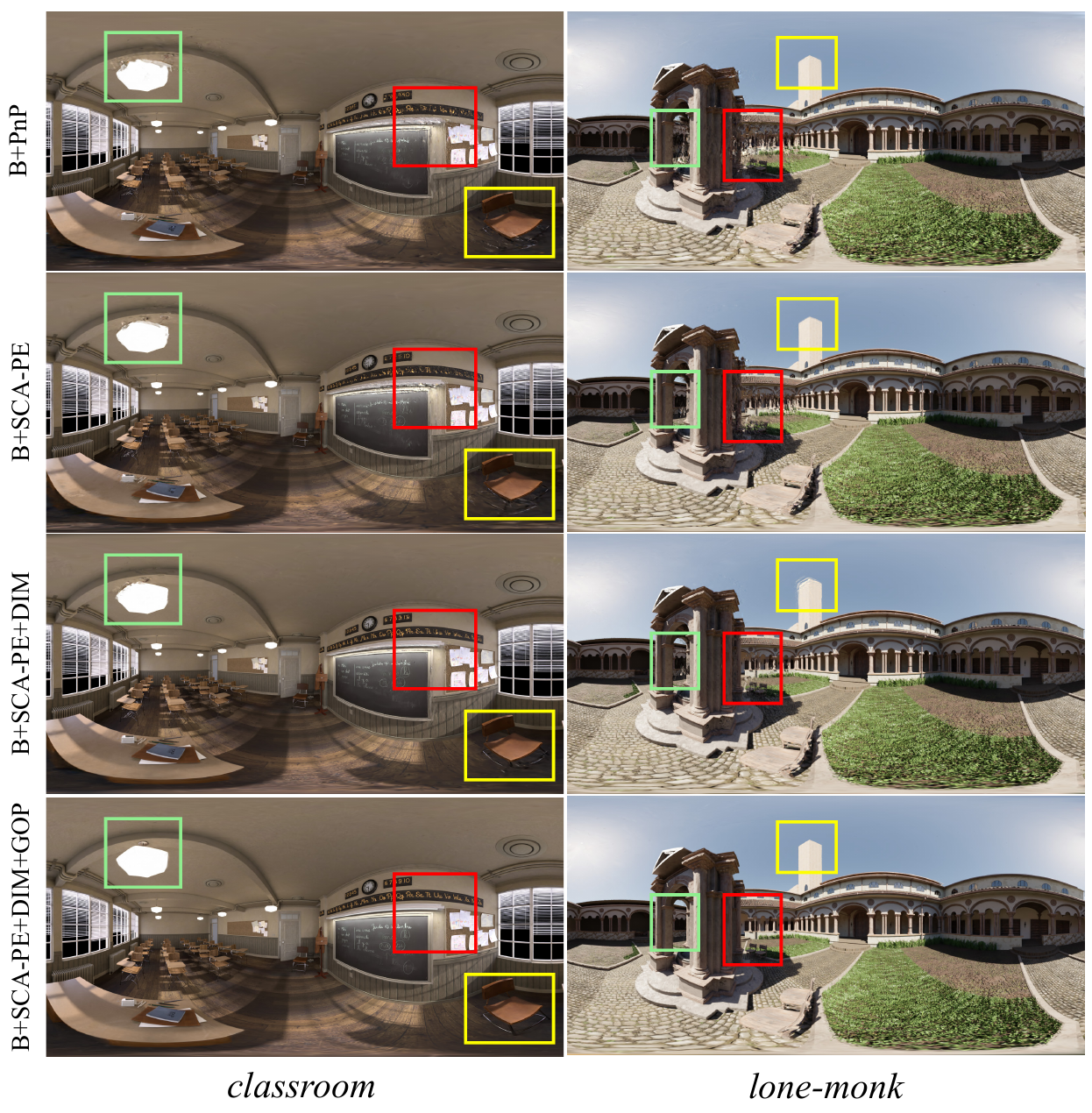}
\caption{\textbf{Visual ablation results of novel view synthesis.}}

\label{fig:4.4-abs-modules}
\end{figure}

As shown in Table~\ref{tab:4.4-abs-modules} and Figure~\ref{fig:4.4-abs-modules}, incorporating the spherical consistency-aware pose estimation (SCA-PE) module (Table~\ref{tab:4.4-abs-modules}, 3rd row, \textit{B+SCA-PE}) into the baseline model (1st row, \textit{B}) significantly improves both novel view synthesis and camera pose estimation.
Furthermore, as indicated in Table~\ref{tab:4.4-abs-modules} (2nd row, \textit{B}+PnP), our method clearly outperforms vanilla PnP-based pose estimation \cite{lin2025longsplat}.
This improvement translates to better novel view synthesis, mitigating artifacts in the \textit{classroom} scene (e.g., lights, blackboard, chair legs) and the \textit{lone-monk} scene (e.g., corridor pillars, highlighted in green and red boxes) as shown in Figure~\ref{fig:4.4-abs-modules}.
When the depth-inlier merging (DIM) module is incorporated (\textit{B+SCA-PE+DIM}), depth cues consistent across multiple views enhance the geometric structure of the Gaussians.
This not only reduces noise in the \textit{classroom} and \textit{lone-monk} (corridor pillars) scenes but also improves the accuracy of camera pose estimation.
Nevertheless, some unreliable Gaussian outliers remain, leading to incomplete distortion removal in the \textit{classroom} and residual artifacts on the white building in \textit{lone-monk}.
Finally, by further introducing the Gaussian outlier pruning (GOP) module (\textit{B+SCA-PE+DIM+GOP}), these artifacts are effectively eliminated, resulting in photorealistic rendered images across all scenes.
These results validate the effectiveness of our model design.

\begin{table}[t]
\caption{\textbf{Ablation experiments on different depth models.}}
\label{tab:4.5-abs-depths}
\centering
\scriptsize
\setlength{\tabcolsep}{3pt} 
\begin{threeparttable}
\begin{tabular}{ccccccc}
\toprule
Method & PSNR & SSIM & LPIPS & RPE\_t & RPE\_r & ATE \\
\midrule
Ours+DepthAnywhere\cite{wang2024depth}   & 29.22 & 0.848 & 0.150 & \underline{\textit{0.389}} & 0.127 & \underline{\textit{0.0129}} \\
Ours+DA$^2$\cite{li20252}         & \underline{\textit{30.61}} & \textbf{0.888} & \underline{\textit{0.123}} & 0.441 & \underline{\textit{0.050}} & 0.0148 \\
Ours+UniK3D\cite{piccinelli2025unik3d}          & \textbf{30.81} & \underline{\textit{0.887}} & \textbf{0.113} & \textbf{0.040} & \textbf{0.016} & \textbf{0.0007}  \\
\bottomrule
\end{tabular}
\end{threeparttable}
\end{table}

\subsubsection*{Ablation Study on Monocular Depth Models.}


Next, we evaluate the performance of our method with different monocular depth models, including DepthAnywhere \cite{wang2024depth}, DA$^2$ \cite{li20252}, and UniK3D \cite{piccinelli2025unik3d}.
As shown in Table~\ref{tab:4.5-abs-depths}, DepthAnywhere and DA$^2$ adopt affine-invariant and scale-invariant losses, respectively, making it challenging to maintain consistent scales between adjacent frames.
Although the depth-inlier merging module aligns the scales and shifts between monocular depthmaps and Gaussian representations, depthmaps from these models still exhibit noticeable adjacent-frame inconsistencies in challenging outdoor environments (e.g., the \textit{fisher-hut} scene in \textbf{Supplementary Materials}), which leads to inaccurate pose reconstruction.
Nevertheless, compared with the methods listed in Tables~\ref{tab:4.1-eval-nvs-ob3d} and \ref{tab:4.3-eval-pose-ob3d}, both DepthAnywhere and DA$^2$ achieve superior novel view synthesis quality and more accurate camera pose estimation.
In contrast, UniK3D produces absolute depth with stronger adjacent-frame consistency, enabling accurate pose reconstruction across all scenes and achieving the best overall performance in novel view synthesis.

\section{Conclusion}
\label{sec:conclusion}


In this work, we propose a pose-free omnidirectional 3DGS framework comprising two key components: a spherical consistency–aware pose estimation module and a depth-inlier–aware densification module.
The spherical consistency–aware pose estimation module extracts reliable 2D–3D correspondences from depth priors with reconstructed Gaussians, enabling accurate camera pose estimation for 360-degree videos.
Subsequently, the depth-inlier–aware densification module aggregates geometrically consistent depth priors from monocular predictions, facilitating efficient frame-by-frame Gaussian densification.
Together, these components enable accurate camera pose estimation and photorealistic panoramic novel view synthesis for 360-degree videos, outperforming existing pose-free and pose-aware methods.

\section*{Acknowledgments}
This work is supported by the National Natural Science Foundation of China (62476262, 62271467, 62306297, 62306296),  the Beijing Natural Science Foundation (4242053, L242096), the Postdoctoral Fellowship Program of CPSF (GZB20250411), and the Fundamental Research Funds for the Central Universities.

{\small
\bibliographystyle{ieeenat_fullname}
\bibliography{11_references}
}

\ifarxiv \clearpage \appendix \input{12_appendix} \fi

\end{document}